\def\year{2022}\relax
\documentclass[letterpaper]{article} 
\usepackage{aaai22}  
\usepackage{times}  
\usepackage{helvet}  
\usepackage{courier}  
\usepackage[hyphens]{url}  
\usepackage{graphicx} 
\usepackage{natbib}  
\usepackage{caption} 
\DeclareCaptionStyle{ruled}{labelfont=normalfont,labelsep=colon,strut=off} 
\usepackage{algorithm}
\usepackage{algorithmic}
\usepackage{algorithm}
\usepackage{algorithmic}
\usepackage{times}
\usepackage{epsfig}
\usepackage{graphicx}
\usepackage{amsmath}
\usepackage{amssymb}
\usepackage{amsmath,amssymb,amsfonts}
\usepackage{algorithm}
\usepackage{booktabs}
\usepackage{algorithmic}
\usepackage{bm}
\usepackage{multirow}
\usepackage{amssymb}
\usepackage{subcaption}
\usepackage{mathtools}
\usepackage{makecell}
\usepackage{enumitem}
\usepackage{caption}
\usepackage{subcaption}
\usepackage{newfloat}
\usepackage{listings}
\floatstyle{ruled}
\newfloat{listing}{tb}{lst}{}
\floatname{listing}{Listing}
\title{Mutual Contrastive Learning for Visual Representation Learning}

\author{
	Chuanguang Yang\textsuperscript{\rm 1,2},
	Zhulin An\textsuperscript{\rm 1}\thanks{Corresponding author.},
	Linhang Cai\textsuperscript{\rm 1,2},
	Yongjun Xu\textsuperscript{\rm 1}
}
\affiliations{
	\textsuperscript{\rm 1}Institute of Computing Technology, Chinese Academy of Sciences, Beijing, China\\
	\textsuperscript{\rm 2}University of Chinese Academy of Sciences, Beijing, China \\
	\{yangchuanguang, anzhulin, cailinhang19g, xyj\}@ict.ac.cn
}

\usepackage{bibentry}

\begin{document}

\maketitle

\begin{abstract}
	We present a collaborative learning method called Mutual Contrastive Learning (MCL) for general visual representation learning. The core idea of MCL is to perform mutual interaction and transfer of contrastive distributions among a cohort of networks. A crucial component of MCL is Interactive Contrastive Learning (ICL). Compared with vanilla contrastive learning, ICL can aggregate cross-network embedding information and maximize the lower bound to the mutual information between two networks. This enables each network to learn extra contrastive knowledge from others, leading to better feature representations for visual recognition tasks. We emphasize that the resulting MCL is conceptually simple yet empirically powerful. It is a generic framework that can be applied to both supervised and self-supervised representation learning. Experimental results on image classification and transfer learning to object detection show that MCL can lead to consistent performance gains, demonstrating that MCL can guide the network to generate better feature representations. Code is available at https://github.com/winycg/MCL.
\end{abstract}

\begin{figure}[t]
	\centering 
	
	\begin{subfigure}[t]{0.4\textwidth}
		\centering
		\includegraphics[width=\textwidth]{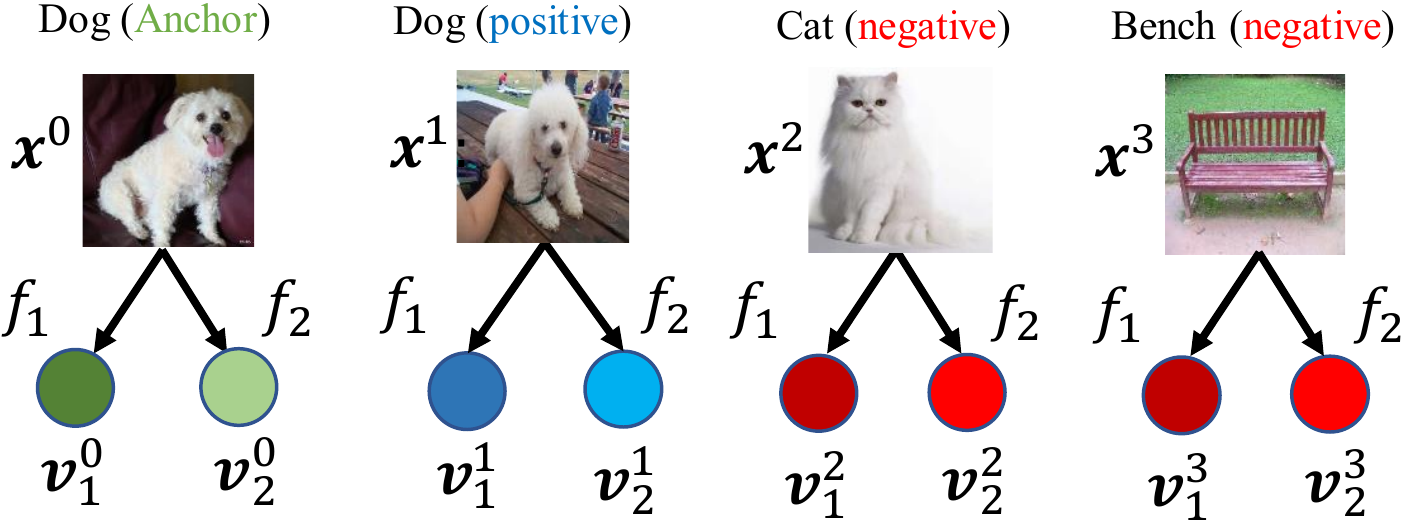}
		\caption{Positive and negative pairs}
		\label{pair_define}
	\end{subfigure}
	\begin{subfigure}[t]{0.4\textwidth}
		\centering
		\includegraphics[width=\textwidth]{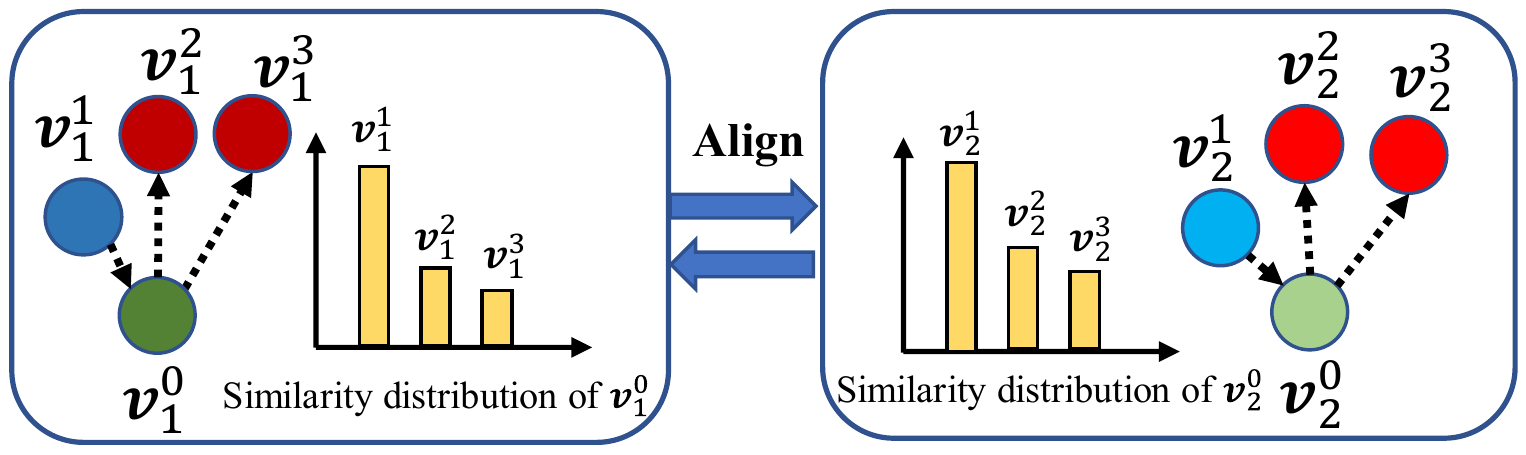}
		\caption{Vanilla Contrastive Learning with Mutual Mimicry}
		\label{scl_}
	\end{subfigure}
	\begin{subfigure}[t]{0.43\textwidth}
		\centering
		\includegraphics[width=\textwidth]{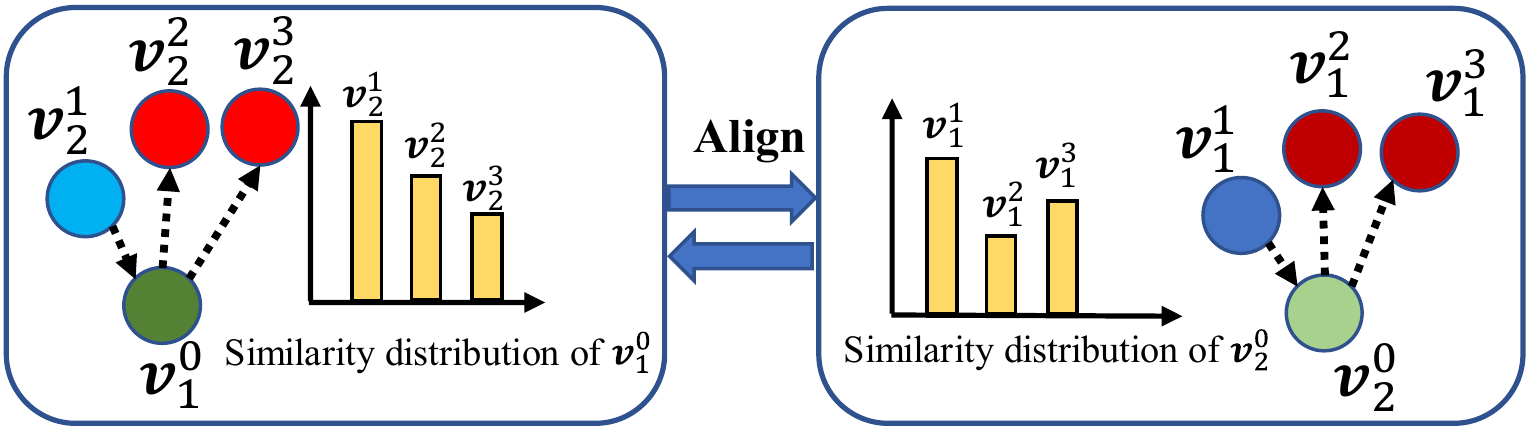}
		\caption{Interactive Contrastive Learning with Mutual Mimicry}
		\label{icl_}
	\end{subfigure}
	\caption{Overview of the proposed \emph{Mutual Contrastive Learning}. $f_{1}$ and $f_{2}$ denote two different networks. $\bm{v}_{m}^{i}$ is the embedding vector inferred from $f_{m}$ with the input sample $\bm{x}^{i}$. The dashed and dotted arrow denotes the direction we want to push close or apart by a contrastive loss. We also perform mutual alignment between two different softmax-based similarity distributions. }
	\label{MCL_pre}
\end{figure} 
\section{Introduction}
Contrastive learning has been widely demonstrated as an effective framework for both supervised~\cite{schroff2015facenet,khosla2020supervised} and self-supervised~\cite{luo2020video, yao2020video,he2020momentum,chen2020simple,li2021dense,zhang2021learning} visual representation learning for artificial intelligence applications~\cite{xu2021artificial}. The core idea of contrastive learning is to pull positive pairs together and push negative pairs apart in the feature embedding space by a contrastive loss. The current pattern of contrastive learning consists of two aspects: (1) how to define positive and negative pairs; (2) how to form a contrastive loss. The main difference between supervised and self-supervised contrastive learning lies in the aspect (1). In the supervised scenario, labels often guide the definition of contrastive pairs. A positive pair is formed by two samples from the same class, while two samples from different classes form a negative pair. In the self-supervised scenario, since we do not have label information, a positive pair is often formed by two views (\emph{e.g.} different data augmentations) of the same sample, while negative pairs are formed by different samples. Given the positive and negative pairs, we can apply a contrastive loss to generate a meaningful feature embedding space. In general, loss functions are independent of how to define pairs. This paper focuses on a generic mutual contrastive loss among multiple networks.

Beyond feature embedding-based contrastive learning, another vein for supervised learning focuses on logit-based learning. The conventional way is to train a network using cross-entropy loss between predictive class probability distribution and the one-hot ground-truth label. Some logit-based online Knowledge Distillation (KD)~\cite{zhang2018deep,zhu2018knowledge,song2018collaborative} methods demonstrate that a cohort of models can benefit from mutual learning of \emph{class probability distributions}. Each model in such a peer-teaching manner learns better compared with learning alone in conventional supervised training. From this perspective, we hypothesize that it may be desirable to perform \emph{mutual contrastive learning} among a cohort of models for learning better \emph{feature representations}. Unlike the class posterior, feature embeddings contain structured knowledge and are more tractable to capture dependencies among various networks. However, existing works~\cite{khosla2020supervised,he2020momentum,chen2020simple} often train a single network to encode data points and apply contrastive learning for its own feature embedding space. Thus it makes sense to take advantage of collaborative learning for better visual representation learning. 

To this end, we propose a simple \emph{Mutual Contrastive Learning} (MCL) framework. The main core of MCL is to perform mutual interaction and transfer of contrastive distributions among a cohort of models. MCL includes Vanilla Contrastive Learning (VCL) and Interactive Contrastive Learning (ICL). Compared with the conventional VCL, our proposed ICL forms contrastive similarity distributions between diverse embedding spaces derived from two different networks. We demonstrate that the objective of ICL is equivalent to maximizing the lower bound to the mutual information between two peer networks. This can be understood to capture dependencies and enable a network to learn extra contrastive knowledge from another network.

Inspired by the idea of DML~\cite{zhang2018deep}, we also perform mutual alignment between different softmax-based contrastive distributions from various networks formed by the same data samples. Similar to DML~\cite{zhang2018deep}, the distributions can be seen as soft labels to supervise others. Such a peer-teaching manner with soft labels takes advantage of representation information embedded in different networks. Over two types of contrastive learning, we can derive \emph{soft VCL label} and \emph{soft ICL label}. Although the soft VCL label has been applied in previous KD works~\cite{ge2020mutual,fang2021seed}, its anchor and contrastive embeddings are still formed from the same network, limiting the information interactions. Instead, our proposed soft ICL label aggregates cross-network embeddings to construct contrastive distributions, which is demonstrated to be more informative than the conventional soft VCL label.  

To maximize the effectiveness of MCL, we summarize VCL and ICL with mutual mimicry into a unified framework, as illustrated in Fig.~\ref{MCL_pre}. MCL helps each model capture extra contrastive knowledge to construct a better representation space. Inspired by DML, since networks start from different initial conditions, each one can learn knowledge that others have not. In fact, MCL can be regarded as a group-wise contrastive loss and is orthogonal to defining positive and negative pairs. Therefore, we can readily apply MCL for both supervised and self-supervised contrastive learning. 

We apply MCL to representation learning for a broad range of visual tasks, including supervised and self-supervised image classification and transfer learning to object detection. MCL can lead to consistent performance gains upon the baseline methods. Note that collaborative learning among a cohort of models is conducted during the training stage. Any network in the cohort can be kept during the inference stage. Compared with the original network, the kept network does not introduce additional inference costs.

Our main contributions are listed as follows: (1) We propose a MCL framework that aims to facilitate mutual interaction and transfer of contrastive knowledge among a cohort of models. (2) MCL is a new collaborative training scheme in terms of representation learning. It is a simple yet powerful framework that can be applied to both supervised and self-supervised representation learning. (3) Thorough experimental results show that MCL can lead to significant performance improvements across frequently-used visual tasks.

\section{Related Work}
\paragraph{Contrastive Learning.} Contrastive learning has been extensively exploited for both supervised and self-supervised visual representation learning. The main idea of contrastive learning is to push positive pairs close and negative pairs apart by a contrastive loss~\cite{hadsell2006dimensionality} to obtain a discriminative space. For supervised learning, contrastive learning is often used for image classification~\cite{khosla2020supervised} and deep metric learning~\cite{schroff2015facenet}. Recently, self-supervised contrastive learning can guide networks to learn general features and achieve state-of-the-art performance for downstream visual recognition tasks. The core idea is to learn invariant representations over human-designed pretext tasks by a contrastive loss~\cite{oord2018representation}. Typical pretext tasks are jigsaw~\cite{noroozi2016unsupervised} used in PIRL~\cite{misra2020self} and data augmentations used in SimCLR~\cite{chen2020simple} and MoCo~\cite{he2020momentum}. This paper does not design a new contrastive method. Instead, our focus is to propose a generic mutual contrastive learning framework. We can incorporate MCL with the above advanced contrastive works to learn better feature representations by taking advantage of collaborative learning. Although previous MVCL~\cite{yang2021multi} also conducts contrastive representation learing with multiple networks, it does not perform explicit transfer of contrastive distributions. Moreover, MVCL can only be applied to the supervised scenario and can not generalize to self-supervised learning. 

\paragraph{Collaborative Learning.} The idea of collaborative learning has been explored in online knowledge distillation. DML~\cite{zhang2018deep} shows that a group of models can benefit from mutual learning of predictive class probability distributions. CL~\cite{song2018collaborative} further extends this idea to a hierarchical architecture with multiple classifier heads. PCL~\cite{wu2021peer} introduces an extra temporal mean network for each peer as the teacher role. HSSAKD~\cite{yang2021knowledge} proposes mutual transfer of self-supervision augmented distributions by extending the teacher-student based counterpart ~\cite{yang2021hierarchical}. In contrast to mutual mimicry, ONE~\cite{zhu2018knowledge}, OKDDip~\cite{chen2020online} and KDCL~\cite{guo2020online} construct an online teacher via a weighted ensemble logit distribution but differ in various aggregation strategies. Beyond logit level, we take advantage of collaborative learning from the perspective of \emph{representation learning}. Moreover, we can readily incorporate MCL with previous logit-based methods together.

\paragraph{Embedding-based Relation Distillation.} Compared with the final class posterior, the latent feature embeddings encapsulates more structural information. Some previous KD methods transfer the embedding-based relational graph where each node represents one sample~\cite{park2019relational,peng2019correlation}. More recently, MMT~\cite{ge2020mutual} employs soft softmax-triplet loss to learn relative similarities from other networks for unsupervised domain adaptation on person Re-ID. To compress networks over self-supervised MoCo~\cite{he2020momentum}, SEED~\cite{fang2021seed} transfers soft InfoNCE-based~\cite{oord2018representation} contrastive distributions from a teacher to a student. A common characteristic of previous works is that contrastive distributions are often constructed from the same embedding space, restricting peer information interactions. Instead, we aggregate cross-network embeddings to model interactive contrastive distributions.

\section{Methodology}
\subsection{Collaborative Learning Architecture}
\textbf{Notation.} A classification network $f(\cdot)$ can be divided into a feature extractor $\varphi(\cdot)$ and a linear FC layer $FC(\cdot)$. $f$ maps an input image $\bm{x}$ to a logit vector $\bm{z}$, \emph{i.e.} $\bm{z}=f(\bm{x})=FC(\varphi(\bm{x}))$.  Moreover, we add an additional projection module $\phi(\cdot)$ that includes two sequential FC layers with a middle ReLU. $\phi(\cdot)$ is to transform a feature embedding from the feature extractor $\varphi(\cdot)$ into a latent embedding $\bm{v}\in \mathbb{R}^{d}$, \emph{i.e.} $\bm{v}=\phi(\varphi(\bm{x}))$, where $d$ is the embedding size. The embedding $\bm{v}$ is used for contrastive learning.

\textbf{Training Graph.} The overall training graph contains $M (M\geqslant 2)$ classification networks denoted by $\{f_{m}\}_{m=1}^{M}$ for collaborative learning. When $M=2$, we use two independent networks $f_{1}$ and $f_{2}$. When $M>2$, the low-level feature layers across $\{f_{m}\}_{m=1}^{M}$ are shared to reduce the training complexity. All the same networks in the cohort are initialized with various weights to learn diverse representations. This is a prerequisite for the success of knowledge mutual learning. Each $f_{m}$ in the cohort is equipped with an additional embedding projection module $\phi_{m}$. The overall training graph is shown in Fig.~\ref{mcl_sup}.

\textbf{Inference Graph.} During the test stage, we discard all projection modules $\{\phi_{m}\}_{m=1}^{M}$ and keep one network for inference. The architecture of the kept network is identical to the original network. That is to say that we do not introduce extra inference costs. Moreover, we can select any $f_{m}$ in the cohort for the final deployment.

\subsection{Mutual Contrastive Learning}
\subsubsection{Vanilla Contrastive Learning (VCL).}
The idea of contrastive loss is to push positive pairs close and negative pairs apart in the latent embedding space. Given an input sample $\bm{x}^{0}$ as the anchor, we can obtain $1$ positive sample $\bm{x}^{1}$ and $K(K\geqslant 1)$ negative samples $\{\bm{x}^{k}\}_{k=2}^{K+1}$. For supervised learning, the positive sample is from the same class with the anchor, while negative samples are from different classes. For self-supervised learning, the anchor and positive samples are often two copies from different augmentations
applied on the same instance, while negative samples are different instances. For ease of notation, we denote the anchor embedding as $\bm{v}_{m}^{0}$, the positive embedding as $\bm{v}_{m}^{1}$ and $K$ negative embeddings as $\{\bm{v}_{m}^{k}\}_{k=2}^{K+1}$. $m$ represents that the embedding is generated from $f_{m}$. Here, feature embeddings are preprocessed by $l_{2}$-normalization. 

We use the dot product to measure similarity distribution between the anchor and contrastive embeddings with \emph{softmax} normalization. Thus, we can obtain contrastive probability distribution $\bm{p}_{m}=softmax([(\bm{v}_{m}^{0}\cdot \bm{v}_{m}^{1}/\tau),(\bm{v}_{m}^{0}\cdot \bm{v}_{m}^{2}/\tau),\cdots,(\bm{v}_{m}^{0}\cdot \bm{v}_{m}^{K+1}/\tau)])$, where $\tau$ is a constant temperature. $\bm{p}_{m}$ measures the relative sample-wise similarities with a normalized probability distribution. A large probability represents a high similarity between the anchor and a contrastive embedding. We use cross-entropy loss to force the positive pair close and negative pairs away upon the contrastive distribution $\bm{p}_{m}$:
\begin{equation}
\mathcal{L}^{VCL}_{m}=-\log{\bm{p}_{m}^{1}}=-\log\frac{\exp(\bm{v}_{m}^{0}\cdot \bm{v}_{m}^{1}/\tau)}{\sum_{k=1}^{K+1}\exp (\bm{v}_{m}^{0}\cdot \bm{v}_{m}^{k}/\tau)}.
\label{scl}
\end{equation}
Here, $\bm{p}_{m}^{k}$ is the $k$-th element of $\bm{p}_{m}$. This loss is equivalent to a $(K+1)$-way softmax-based classification loss that forces the network to classify the positive sample correctly. In fact, the form of Eq.(\ref{scl}) is an InfoNCE loss~\cite{oord2018representation}, which has been widely used in recent self-supervised contrastive learning~\cite{he2020momentum}. When applying contrastive learning to a cohort of $M$ networks, the vanilla method is to summarize each contrastive loss:
\begin{equation}
\mathcal{L}^{VCL}_{1\sim M}=\sum_{m=1}^{M}(\mathcal{L}_{m}^{VCL}).
\end{equation}

\subsubsection{Interactive Contrastive Learning (ICL).}
However, vanilla contrastive learning does not model cross-network relationships for collaborative learning. This is because the contrastive distribution is learned from the network's own embedding space. To take full advantage of information interaction among various peer networks, we propose a novel \emph{Interactive Contrastive Learning} (ICL) to model cross-network interactions to learn better feature representations. We formulate ICL for the case of two parallel networks $f_{a}$ and $f_{b}$, where $a,b\in \{1,2,\cdots,M\},a\neq b$, and then further extend ICL to more than two networks among $\{f_{m}\}_{m=1}^{M}$.

To conduct ICL, we first fix $f_{a}$ and enumerate over $f_{b}$. Given the anchor embedding $\bm{v}_{a}^{0}$ extracted from $f_{a}$, we enumerate the positive embedding $\bm{v}_{b}^{1}$ and negative embeddings  $\{\bm{v}_{b}^{k}\}_{k=2}^{K+1}$ extracted from $f_{b}$. Here, both $\{\bm{v}_{a}^{k}\}_{k=0}^{K+1}$ and $\{\bm{v}_{b}^{k}\}_{k=0}^{K+1}$ are generated from the same $K+1$ samples $\{\bm{x}^{k}\}_{k=0}^{K+1}$ correspondingly, as illustrated in Fig.~\ref{pair_define}. The contrastive probability distribution from $f_{a}$ to $f_{b}$ can be formulated as $\bm{q}_{a\rightarrow  b}=softmax([(\bm{v}_{a}^{0}\cdot \bm{v}_{b}^{1}/\tau),(\bm{v}_{a}^{0}\cdot \bm{v}_{b}^{2}/\tau),\cdots,(\bm{v}_{a}^{0}\cdot \bm{v}_{b}^{K+1}/\tau)])$. Similar to Eq.(\ref{scl}), we use cross-entropy loss upon the contrastive distribution $\bm{q}_{a\rightarrow  b}$:
\begin{align}
\mathcal{L}^{ICL}_{a\rightarrow  b}&=-\log{\bm{q}_{a\rightarrow  b}^{1}}
=-\log\frac{\exp (\bm{v}_{a}^{0}\cdot \bm{v}_{b}^{1}/\tau)}{\sum_{k=1}^{K+1}\exp (\bm{v}_{a}^{0}\cdot \bm{v}_{b}^{k}/\tau)}.
\label{icl_hard}
\end{align}
Here, $\bm{q}_{a\rightarrow  b}^{k}$ is the $k$-th element of $\bm{q}_{a\rightarrow  b}$. We can observe that the main difference between Eq.(\ref{scl}) and Eq.(\ref{icl_hard}) lies in various types of embedding space for generating contrastive distributions. Compared with Eq.(\ref{scl}), Eq.(\ref{icl_hard}) employs contrastive embeddings from another network instead of the network's own embedding space. It can model explicit corrections or dependencies in various embedding spaces among multiple peer networks, facilitating  information communications to learn better feature representations.

\emph{Theoretical Analysis.} Compared to Eq.(\ref{scl}), we attribute the superiority of minimizing Eq.(\ref{icl_hard}) to maximizing the lower bound on the mutual information $I(\bm{v}_{a}^{0},\bm{v}_{b}^{1})$ between $f_{a}$ and $f_{b}$, which is formulated as:
\begin{equation}
I(\bm{v}_{a}^{0},\bm{v}_{b}^{1})\geq \log(K)- \mathbb{E}_{(\bm{v}_{a}^{0},\bm{v}_{b}^{1})}\mathcal{L}^{ICL}_{a\rightarrow  b}
\label{mutual_infor}.
\end{equation}

Inspired by~\cite{tian2019contrastive}, detailed proof from Eq.(\ref{icl_hard}) to derive Eq.(\ref{mutual_infor}) is provided in Appendix. Intuitively, the mutual information $I(\bm{v}_{a}^{0},\bm{v}_{b}^{1})$ measures the reduction of uncertainty in contrastive feature embeddings from $f_{b}$ when the anchor embedding from $f_{a}$ is known. This can be understood that each network could gain extra contrastive knowledge from others benefiting from Eq.(\ref{icl_hard}). Thus, it can lead to better representation learning than independent contrastive learning of Eq.(\ref{scl}). As $K$ increases, the mutual information $I(\bm{v}_{a}^{0},\bm{v}_{b}^{1})$ would be higher, indicating that $f_{a}$ and $f_{b}$ could learn more common knowledge from each other. 

When extending to $\{f_{m}\}_{m=1}^{M}$, we perform ICL in every two of $M$ networks to model fully connected dependencies, leading to the overall loss as: 
\begin{equation}
\mathcal{L}_{1\sim M}^{ICL}=\sum_{1\leq a<b\leq M}^{M}(\mathcal{L}^{ICL}_{a\rightarrow  b}+\mathcal{L}^{ICL}_{b\rightarrow  a}) 
\end{equation}

\subsubsection{Soft Contrastive Learning with Online Mutual Mimicry}
The success of \emph{Deep Mutual Learning}~\cite{zhang2018deep} suggests that each network can generalize better from mutually learning other networks' soft class probability distributions in an online peer-teaching manner. This is because the output of class posterior from each network can be seen as a natural \emph{soft label} to supervise others. Based on this idea, it is desirable to derive soft contrastive distributions as \emph{soft labels} from contrastive learning, for example, $\bm{p}_{m}$ from VCL and $\bm{q}_{a\rightarrow  b}$ from ICL. In theory, both $\bm{p}_{m}$ and $\bm{q}_{a\rightarrow  b}$ can also be seen as class posteriors. Thus it is theoretically reasonable to perform mutual mimicry of these contrastive distributions for better representation learning.

Specifically, we utilize Kullback Leibler (KL)-divergence to force each network's contrastive distributions to align corresponding soft labels provided from other networks within the cohort. This paper focuses on mutually mimicking two types of contrastive distributions from VCL and ICL:

	\textbf{\emph{Soft Vanilla Contrastive Learning (Soft VCL).}} For refining $\bm{p}_{m}$ from $f_{m}$, the soft pseudo labels are peer contrastive distributions $\{\bm{p}_{l}\}_{l=1,l\neq m }^{l=M}$ generated from $\{f_{l}\}_{l=1,l\neq m }^{l=M}$, respectively. We use $\mathbf{KL}$ divergence to force $\bm{p}_{m}$ to align them. For applying soft VCL to the cohort of $\{f_{m}\}_{m=1}^{M}$, the overall loss can be formulated as:
	\begin{equation}
	\mathcal{L}_{1\sim M}^{Soft\_VCL}=\sum_{m=1}^{M}\sum_{l=1,l\neq m }^{M}\mathbf{KL
	}(\bm{p}_{l}\parallel \bm{p}_{m}).
	\label{VSCL}
	\end{equation}
	Here, $\bm{p}_{l}$ is the soft label detached from gradient back-propagation for stability.
	
	\textbf{\emph{Soft Interactive Contrastive Learning (Soft ICL).}} Given two networks $f_{a}$ and $f_{b}$, we can derive interactive contrastive distributions $\bm{q}_{a\rightarrow b}$ and $ \bm{q}_{b\rightarrow a}$ using ICL. It makes sense to force the consistency between $\bm{q}_{a\rightarrow b}$ and $ \bm{q}_{b\rightarrow a}$ for mutual calibration by Soft ICL. When extending to $\{f_{m}\}_{m=1}^{M}$, we perform Soft ICL in every two of $M$ networks, leading to the overall loss as: 
	\begin{equation}
	\mathcal{L}_{1\sim M}^{Soft\_ICL}=\sum_{a=1}^{M}\sum_{b=1,b\neq a }^{M}\mathbf{KL
	}(\bm{q}_{b\rightarrow a}\parallel \bm{q}_{a\rightarrow b}).
	\label{VICL}
	\end{equation}
	Here, $\bm{q}_{b\rightarrow a}$ is the soft label detached from gradient back-propagation for stability.

\subsubsection{Discussion with Soft VCL and Soft ICL.} We remark that using a vanilla contrastive distribution $\bm{p}$ as a soft label has been explored by some previous works~\cite{ge2020mutual,fang2021seed}. These works often construct contrastive relationships using embeddings from the same network, as illustrated in Eq.(\ref{scl}). In contrast, we propose an interactive contrastive distribution $\bm{q}$ to perform Soft ICL. Intuitively, $\bm{q}$ aggregates cross-network embeddings to model the soft label, which is more informative than $\bm{p}$ constructed from a single embedding space. Moreover, refining a better $\bm{q}$ may decrease $\mathcal{L}^{ICL}_{a\rightarrow  b}$, further maximizing the lower bound on the mutual information $I(\bm{v}_{a}^{0}, \bm{v}_{b}^{1})$ between $f_{a}$ and $f_{b}$. Compared with soft VCL, soft ICL can facilitate more adequate interactions among multiple networks. Empirically, we found soft ICL excavates better performance gains by taking full advantage of collaborative contrastive learning.

\begin{figure*}[htbp]  
	\centering
	\includegraphics[width=0.9\linewidth]{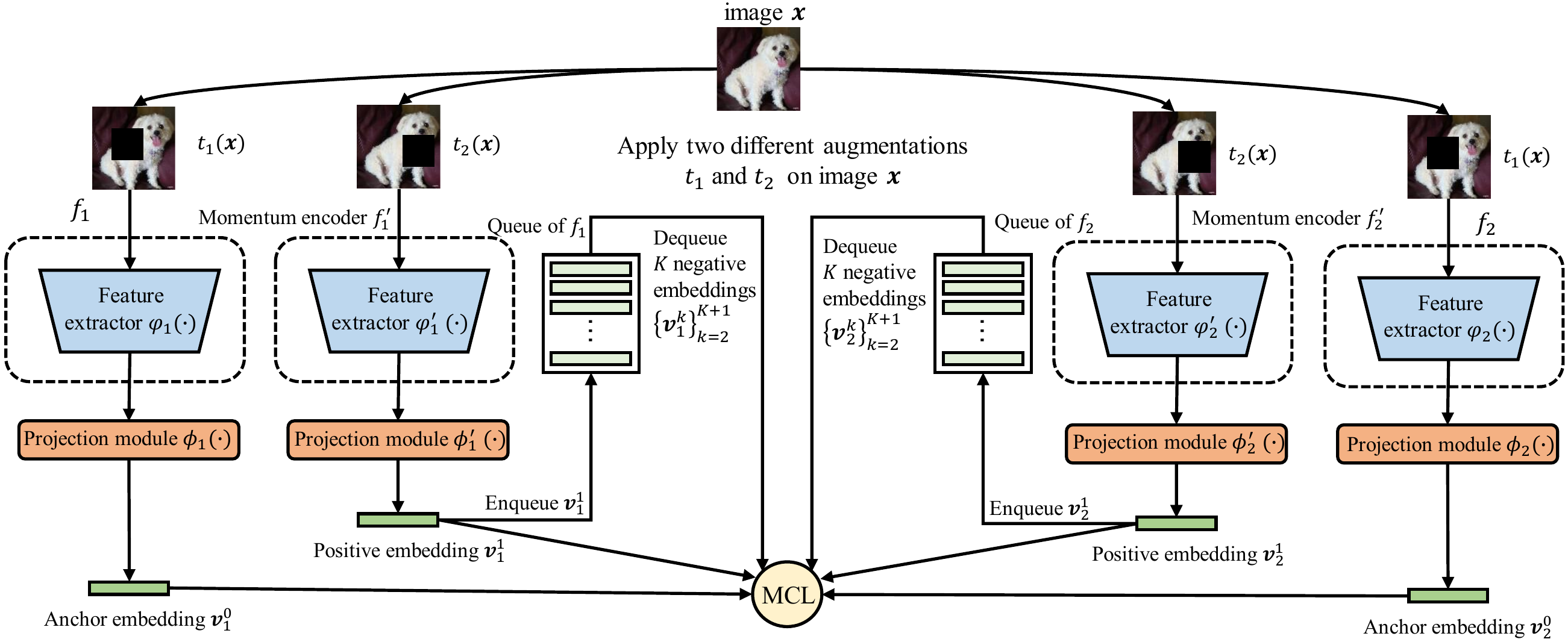}
	\caption{Overview of incorporate MCL with MoCo~\cite{he2020momentum} for self-supervised learning.}
	\label{moco_mcl}
\end{figure*}
\subsubsection{Overall loss of MCL.}
To take full advantage of collaborative learning, we summarize all contrastive loss terms as the overall loss for MCL among a cohort of $M$ networks:

\begin{align}
\mathcal{L}^{MCL}_{1\sim M}=&\alpha\mathcal{L}^{VCL}_{1\sim M}+\beta \mathcal{L}^{ICL}_{1\sim M} \notag \\
&+\gamma \mathcal{L}^{Soft\_VCL}_{1\sim M}+\lambda \mathcal{L}^{Soft\_ICL}_{1\sim M},
\label{MCL}
\end{align}

where $\alpha$, $\beta$, $\gamma$ and $\lambda$ are weight coefficients. We set $\alpha=\beta=0.1$ in supervised learning and $\alpha=\beta=1$ in self-supervised learning. Moreover, we set $\gamma=\lambda=1$ for KL-divergence losses.

\begin{figure}[htbp]  
	\centering
	\includegraphics[width=1\linewidth]{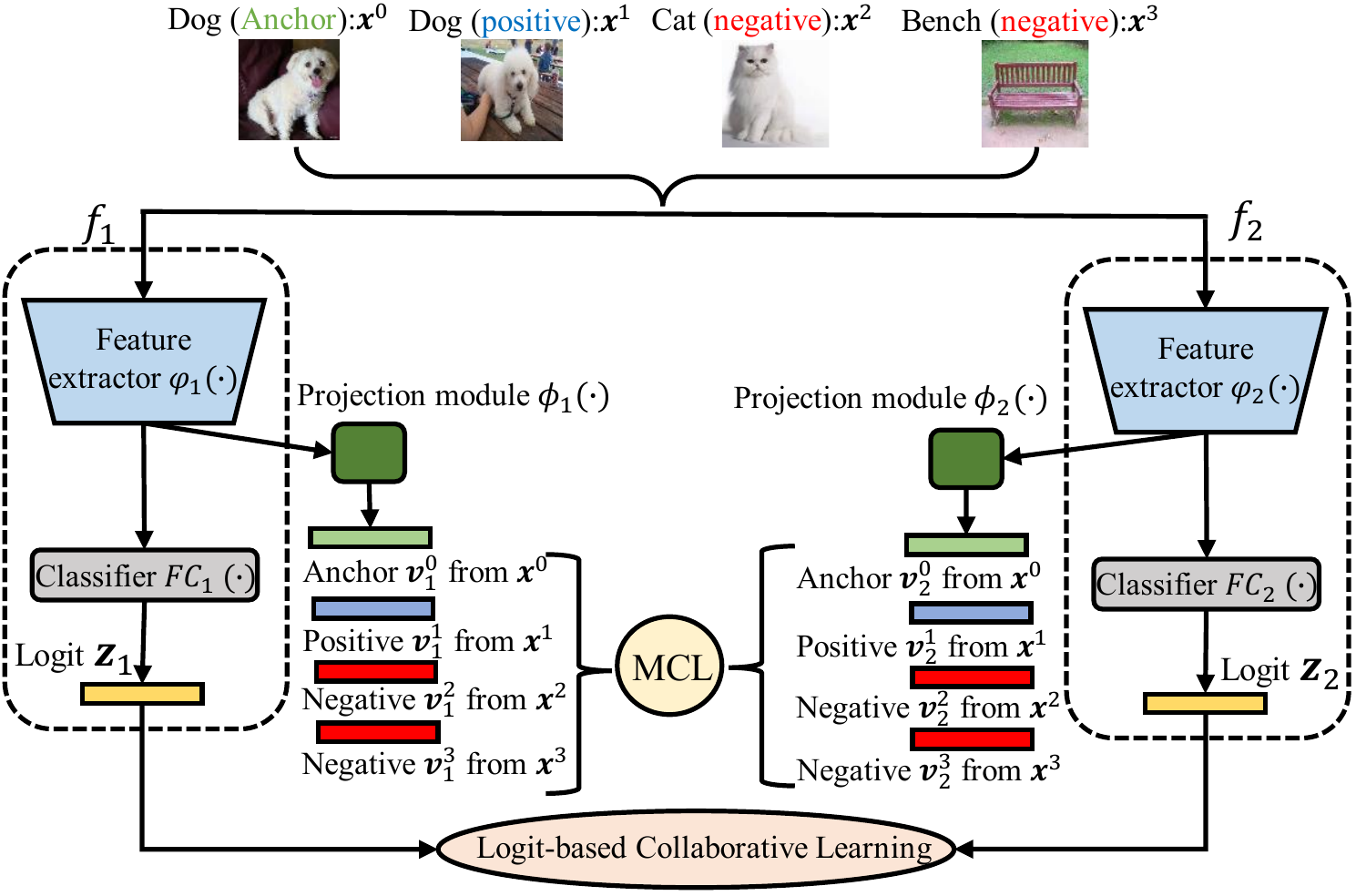}
	\caption{Overview of MCL for supervised learning.}
	\label{mcl_sup}
\end{figure}

\subsection{Apply MCL to Supervised Learning}
For the small-scale dataset like CIFAR-100, we create a class-aware sampler to derive contrastive samples from the mini-batch. The mini-batch with a batch size of $B$ consists of $B/2$ classes. Each class has two samples, and others from different classes are negative samples. We regard each sample as an anchor instance and others as contrastive instances within the current mini-batch. For the large-scale dataset like ImageNet, we create an online memory bank~\cite{wu2018unsupervised} to store massive embeddings since the batch size limits the number of available contrastive samples.

For supervised learning, we also perform the conventional sample-independent logit-based learning. For $M$ networks $\{f_{m}\}_{m=1}^{M}$ with the input $\bm{x}$, their generated logit vectors are $\{\bm{z}_{m}\}_{m=1}^{M}$. Each network is supervised by a cross-entropy loss $\mathcal{L}_{ce}$ between the predictive probability distribution and the ground-truth label $y$. The total loss is:
$
\mathcal{L}^{vanilla}_{1\sim M}=\sum_{m=1}^{M}\mathcal{L}_{ce}(softmax(\bm{z}_{m}),y)
\label{vanilla}$.

We summarize the logit-based classification loss and embedding-based MCL as the overall loss for collaborative learning. The overall loss is $\mathcal{L}^{sup}_{1\sim M}$:
\begin{equation}
\label{overall_loss}
\mathcal{L}^{sup}_{1\sim M}=\mathcal{L}^{vanilla}_{1\sim M}+\mathcal{L}^{MCL}_{1\sim M}
\end{equation}
We illustrate the overview of collaborative learning under the supervised scenario in Fig.~\ref{mcl_sup}.

\begin{table*}[tbp]
	\centering
	\resizebox{0.95\linewidth}{!}{
		\begin{tabular}{c|cccccc|cc}  
			\toprule
			Network&Baseline&DML&CL&ONE&OKDDip&MVCL&MCL($\times 4$)+Logit &Gain ($\uparrow$)\\ 
			\midrule
			
			WRN-16-2&72.55$_{\pm 0.24}$&75.04$_{\pm 0.22}$&74.18$_{\pm 0.34}$&74.04$_{\pm 0.19}$&74.99$_{\pm 0.45}$&\underline{75.76}$_{\pm 0.21}$&\textbf{76.34}$_{\pm 0.22}$&0.58\\
			
			WRN-40-2&76.89$_{\pm 0.29}$&78.45$_{\pm 0.42}$&78.64$_{\pm 0.31}$&79.05$_{\pm 0.22}$&\underline{79.21}$_{\pm 0.06}$&79.16$_{\pm 0.36}$&\textbf{80.02}$_{\pm 0.45}$&0.81\\
			
			WRN-28-4&79.17$_{\pm 0.29}$&80.54$_{\pm 0.38}$&80.83$_{\pm 0.27}$&80.58$_{\pm 0.17}$&80.47$_{\pm 0.27}$&\underline{81.16}$_{\pm 0.36}$&\textbf{81.68}$_{\pm 0.31}$ &0.52 \\
			
			ShuffleNetV2 1$\times$&70.93$_{\pm 0.24}$&75.35$_{\pm 0.30}$&\underline{75.94}$_{\pm 0.25}$&75.74$_{\pm 0.33}$&75.24$_{\pm 0.30}$&75.88$_{\pm 0.13}$&\textbf{77.02}$_{\pm 0.32}$ &1.08 \\
			
			HCGNet-A2&79.00$_{\pm 0.41}$&\underline{82.10}$_{\pm 0.29}$&81.94$_{\pm 0.11}$&80.64$_{\pm 0.20}$&80.11$_{\pm 0.19}$&82.04$_{\pm 0.15}$&\textbf{82.47}$_{\pm 0.20}$ &0.37 \\
			\bottomrule
	\end{tabular}}
	\caption{Top-1 accuracy (\%) of jointly training \emph{four networks} with the same architecture on CIFAR-100. The bold number represents the best result among various methods. 'Gain' indicates the accuracy improvement of MCL upon \underline{the second-best result}.}
	\label{cifar} 
\end{table*}

\begin{table*}[tbp]
	\centering
	\resizebox{0.95\linewidth}{!}{
		\begin{tabular}{c|ccccccc|cc}  
			\toprule
			Network&Baseline&DML&CL&ONE&OKDDip&PCL&MVCL&MCL($\times 3$)+Logit &Gain ($\uparrow$)\\
			\midrule 
			ResNet-18&69.76&69.82$_{\pm 0.08}$&70.04$_{\pm 0.05}$&70.18$_{\pm 0.13}$&69.93$_{\pm 0.06}$&70.42$_{\pm 0.13}$&\underline{70.46}$_{\pm 0.09}$&\textbf{70.82}$_{\pm 0.06}$&0.36\\
			\bottomrule
	\end{tabular}}
	\caption{Top-1 accuracy (\%) of jointly training \emph{three networks} with the same architecture on ImageNet. Part of compared results are obtained from PCL~\cite{wu2021peer}. 'Logit' represents logit-based collaborative learning~\cite{yang2021multi}.}
	\label{imagenet} 
\end{table*}

\subsection{MCL on Self-Supervised Learning}
When MCL is used for self-supervised learning, a positive pair includes two copies from different augmentations applied on the same sample, while negative pairs are often constructed from different samples. Because recent contrastive-based self-supervised learning often uses an InfoNCE loss, \emph{i.e.} the form of Eq.(\ref{scl}), MCL can be readily incorporated with those great works, \emph{e.g.} MoCo~\cite{he2020momentum}. As shown in Fig.~\ref{moco_mcl}, we illustrate the overview of how to incorporate MCL with MoCo for learning visual representation among a cohort of models. Because self-supervised learning often needs a large number of negative samples, MoCo constructs a momentum encoder and a queue for providing contrastive embeddings. Self-supervised learning only involves feature embedding-based learning so that the overall loss is formulated as the MCL loss.

\section{Experiments}

\subsection{Supervised Image Classification}
\textbf{Datasets.} We use CIFAR-100~\cite{krizhevsky2009learning} and ImageNet~\cite{deng2009imagenet} datasets for image classification, following the standard data augmentation and preprocessing pipeline~\cite{huang2017densely}.

\textbf{Hyper-parameters settings.} Following SimCLR~\cite{chen2020simple}, we use $\tau=0.1$ on CIFAR and  $\tau=0.07$ on ImageNet for similarity calibration of $\mathcal{L}^{VCL}$ and $\mathcal{L}^{ICL}$. The contrastive embedding size $d$ is $128$. In soft losses of $\mathcal{L}^{Soft\_VCL}$ and $\mathcal{L}^{Soft\_ICL}$, we utilize $\tau=0.1\times 3=0.3$ on CIFAR and $\tau=0.07\times 3=0.21$ on ImageNet to smooth similarity distributions. For CIFAR-100, we use $K=126$ as the number of negative samples due to a batch size of $128$. For ImageNet, we retrieve one positive and $K=8192$ negative embeddings from the memory bank.

\textbf{Training settings.} For CIFAR-100, all networks are
trained by SGD with a momentum of 0.9, a batch size of 128 and a weight decay of $5\times 10^{-4}$. We use a cosine learning rate that starts from 0.1 and gradually decreases to 0 throughout the 300 epochs. For ImageNet, all networks are trained by SGD with a momentum of 0.9, a batch size of 256 and a weight decay of
$1\times 10^{-4}$. The initial learning rate starts at 0.1 and
is decayed by a factor of 10 at 30 and 60 epochs within the total 90 epochs. We conduct all experiments with the same training settings and report the mean result over three runs for a fair comparison.

\begin{table}[t]
	\centering 
	\resizebox{1\linewidth}{!}{
		\begin{tabular}{c|c|cc|cc}  
			\toprule
			Network&Baseline&MCL($\times 2$)&Gain ($\uparrow$)&MCL($\times 4$)&Gain ($\uparrow$) \\  
			\midrule
			ResNet-32&70.91$_{\pm 0.14}$&72.96$_{\pm 0.28}$&2.05&74.04$_{\pm 0.07}$&3.13\\
			
			ResNet-56&73.15$_{\pm 0.23}$&74.48$_{\pm 0.23}$&1.33&75.74$_{\pm 0.16}$&2.59 \\
			
			ResNet-110&75.29$_{\pm 0.16}$&77.12$_{\pm 0.20}$&1.83&78.82$_{\pm 0.14}$&3.53\\
			
			WRN-16-2&72.55$_{\pm 0.24}$&74.56$_{\pm 0.11}$&2.01&75.79$_{\pm 0.07}$&3.24 \\
			
			WRN-40-2&76.89$_{\pm 0.29}$&77.51$_{\pm 0.42}$&0.62&78.84$_{\pm 0.22}$&1.95 \\
			
			HCGNet-A1&77.42$_{\pm 0.16}$&78.62$_{\pm 0.26}$&1.20&79.50$_{\pm 0.15}$&2.08 \\
			
			ShuffleNetV2 0.5$\times$&67.39$_{\pm 0.35}$&69.55$_{\pm 0.22}$&2.16&70.92$_{\pm 0.28}$&3.53 \\
			
			ShuffleNetV2 1$\times$&70.93$_{\pm 0.24}$&73.26$_{\pm 0.18}$&2.33&75.18$_{\pm 0.25}$&4.25 \\
			
			\bottomrule			
	\end{tabular}	}
	\caption{Top-1 accuracy (\%) of jointly training \emph{two or four networks} with the same architecture on CIFAR-100. $\times M$ indicates the cohort has $M$ networks for MCL. 'Gain' indicates the accuracy improvement of MCL upon the Baseline. }
	\label{same_arch} 
\end{table}
\begin{table}[tbp]
	\centering
	
	\resizebox{1.\linewidth}{!}{
		\begin{tabular}{l|cc|cc}  
			\toprule 
			
			\multirow{2}{*}{Method}&\multicolumn{2}{c|}{ResNet-18} &\multicolumn{2}{c}{ResNet-34} \\  
			&ImageNet & Pascal VOC&ImageNet & Pascal VOC \\  
			\midrule
			Baseline&69.76&76.18&73.30&79.81 \\
			MCL ($\times 2$)&70.32 $(\uparrow 0.56)$&77.20 $(\uparrow 1.02)$&74.13 $(\uparrow 0.83)$&80.37 $(\uparrow 0.56)$ \\
			MCL ($\times 4$)&\textbf{70.77} $(\uparrow 1.01)$&\textbf{77.68} $(\uparrow 1.50)$&\textbf{74.34} $(\uparrow 1.04)$&\textbf{80.81} $(\uparrow 1.00)$ \\
			\bottomrule
	\end{tabular}}
	\caption{Top-1 classification accuracy (\%) on ImageNet by jointly training \emph{two or four networks} and mAP(\%) of downstream transfer learning to object detection on Pascal VOC over Faster-RCNN~\cite{ren2016faster} framework. The
		number in brackets is the gain upon Baseline.}
	\label{Detection}
\end{table}

\paragraph{Apply MCL upon baseline on CIFAR-100.}
As shown in Table~\ref{same_arch}, we first investigate the efficacy of MCL upon the conventional supervised training. We apply widely used ResNets~\cite{he2016deep},
WRNs~\cite{zagoruyko2016wide}, HCGNets~\cite{yang2020gated} and ShuffleNetV2~\cite{ma2018shufflenet} as the backbone networks to evaluate the performance.  All results are achieved from jointly training two networks by MCL($\times 2$) or four networks by MCL($\times 4$) with the same architecture. We observe that our MCL($\times 2$) leads to an average improvement of 1.69\% across various architectures upon the independent training for an individual network. The results indicate that MCL can help each network learn better representations effectively. When extending MCL($\times 2$) to MCL($\times 4$), the accuracy gains get more significant. MCL($\times 4$) further advances an average improvement of 3.04\% upon baseline. These results verify our claim that more networks in the cohort can capture richer contrastive knowledge, conducive to
representation learning.

\paragraph{Training complexity on CIFAR-100.} We examine training costs introduced by MCL. For independently training two networks with the same architecture by the conventional cross-entropy loss, the training time and GPU memory are $2\times$ than one network. MCL needs to compute similarity distributions with contrastive embeddings. For supervised contrastive learning on MCL($\times 2$), we use $128$-d embedding size, $1$ positive and $126$ negative embeddings. Extra computation is $4\times (1+126)\times 128\approx 0.07$M FLOPs for each sample, where $4$ represents $2$ VCL and $2$ ICL distributions. Given two ResNet-110 with 335M FLOPs for an example, applying MCL only introduces an extra $0.02\%$ computation. Since MCL derives contrastive embeddings in mini-batch, we did not find a distinct change of GPU memory cost.

\paragraph{Apply MCL upon baseline on ImageNet.} Extensive experiments on more challenging ImageNet further show the scalability of MCL for representation learning to the large-scale dataset. As shown in Table~\ref{Detection}, MCL leads to consistent performance improvements over top-1 and top-5 accuracy. 

\paragraph{Transferring features to object detection.} We use pre-trained ResNet-18 on ImageNet as the backbone over Faster-RCNN~\cite{ren2016faster} for downstream object detection on Pascal VOC~\cite{everingham2010pascal}. The model is finetuned on \texttt{trainval07+12} and evaluated on \texttt{test2007} using mAP. The fine-tuning strategy follows the original implementation~\cite{ren2016faster}. As shown in Table~\ref{Detection}, using MCL for training feature extractors of ResNet-18 and ResNet-34 on ImageNet achieves significant mAP gains consistently for downstream detection. The results demonstrate the efficacy of MCL for learning better representations to downstream semantic recognition tasks.

\paragraph{Comparison with SOTAs.} 
We compare MCL with recent collaborative learning methods, including DML~\cite{zhang2018deep}, CL~\cite{song2018collaborative}, ONE~\cite{zhu2018knowledge}, OKDDip~\cite{chen2020online}, PCL~\cite{wu2021peer} and MVCL~\cite{yang2021multi}. To maximize performance gains, we also incorporate MCL with logit-based collaborative learning~\cite{yang2021multi} to distill class posterior information. As shown in Table~\ref{cifar} and~\ref{imagenet}, our MCL achieves the best performance gains against prior works across various networks. It surpasses the previous SOTA MVCL by an average margin of 0.67\% on CIFAR-100 and a margin of 0.36\% over ResNet-18 on ImageNet. Moreover, it is hard to say which is the second-best method since different methods are superior for various architectures or datasets. These results demonstrate that exploring contrastive representation may be an effective way for collaborative learning beyond class posterior.

\begin{figure}
	\centering 
	
	\begin{subfigure}[t]{0.48\textwidth}
		\centering
		\includegraphics[width=\textwidth]{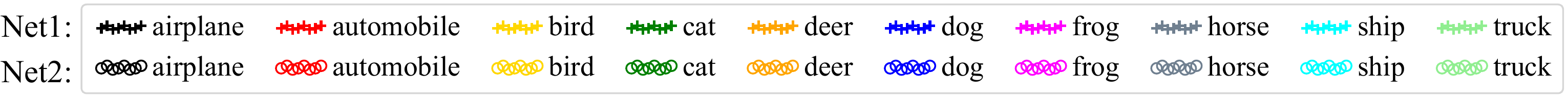}
		\caption{Legend of object classes from CIFAR-10.}
		\label{classes}
	\end{subfigure}
	\begin{subfigure}[t]{0.23\textwidth}
		\centering
		\includegraphics[width=\textwidth]{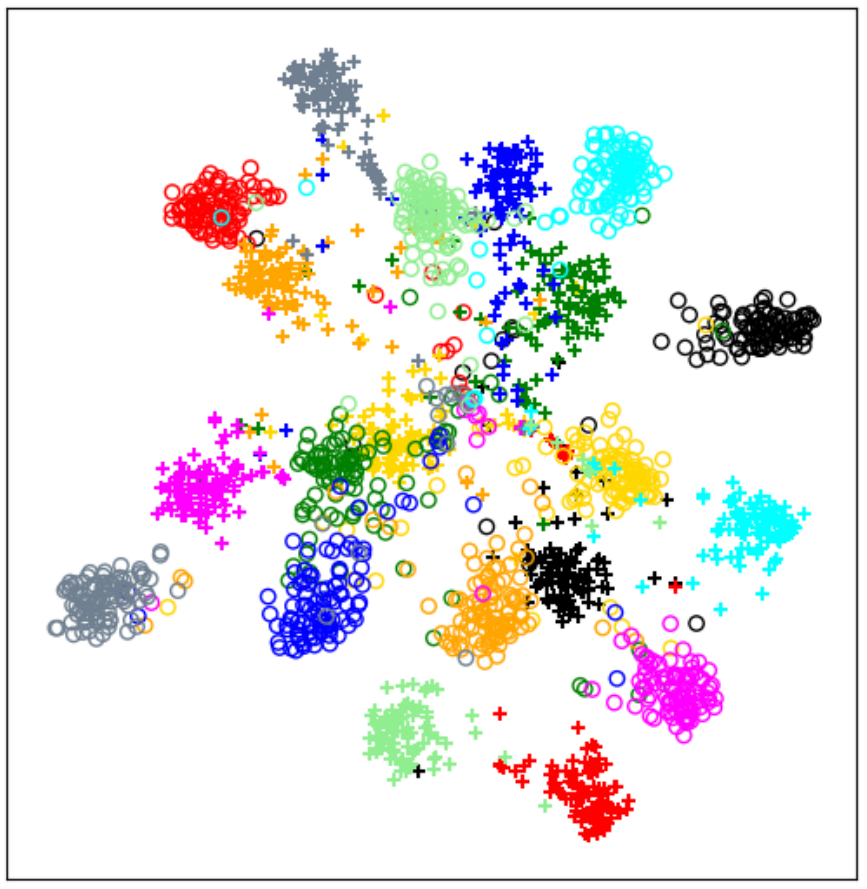}
		\caption{Independent training.}
		\label{Independent}
	\end{subfigure}
	\begin{subfigure}[t]{0.23\textwidth}
		\centering
		\includegraphics[width=\textwidth]{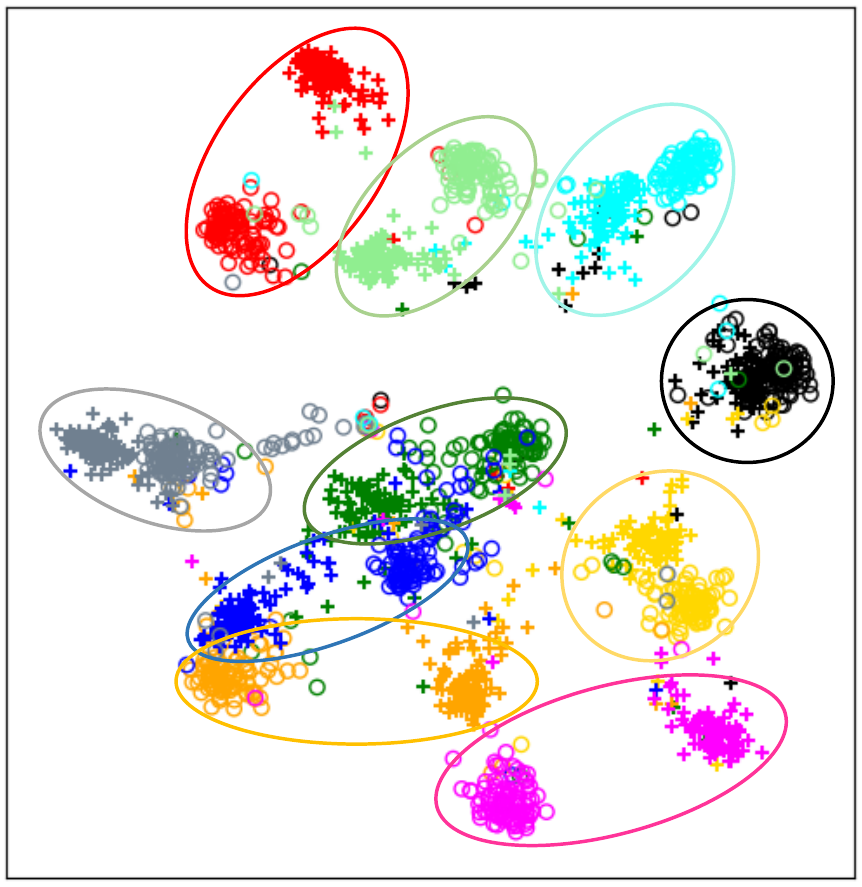}
		\caption{Our proposed MCL.}
		\label{mcl_tsne}
	\end{subfigure}
	\caption{T-SNE visualization of embedding spaces for two ResNet-32 (Net1 and Net2) with independent training (\emph{left}) and our MCL (\emph{right}) on CIFAR-10 dataset. The clusters in the same circle are from the same class.} 
	\label{tsne}
\end{figure} 

\begin{table}[t]
	\centering
	\resizebox{1.\linewidth}{!}{
		\begin{tabular}{l|cc|cc}  
			\toprule
			Network&MoCo&MCL($\times 2$)&MoCoV2&MCL($\times 2$)\\ 
			\midrule
			ResNet-18 & 47.45$_{\pm 0.11}$ & 48.04$_{\pm 0.13}$& 52.30$_{\pm 0.09}$& 52.76$_{\pm 0.06}$ \\
			\bottomrule
	\end{tabular}}
	\caption{Top-1 accuracy (\%) for self-supervised contrastive learning on ImageNet.}
	\label{moco_mcl_exp}
\end{table}

\subsection{Apply MCL to Self-Supervised Learning}
We incorporate MCL with recent self-supervised contrastive learning methods of MoCo~\cite{he2020momentum} and MoCoV2~\cite{chen2020improved}. We follow the standard experimental settings and linear classification protocol~\cite{he2020momentum}. As illustrated in Fig.~\ref{moco_mcl}, we use two networks of $f_{1}$ and $f_{2}$ with two peer momentum encoders of $f^{'}_{1}$ and $f^{'}_{2}$ respectively. As shown in Table~\ref{moco_mcl_exp}, MCL improves popular MoCo and MoCoV2 with 0.59\% and 0.46\%  accuracy improvements on ImageNet, respectively. The results indicate that MCL can help these methods to learn better self-supervised feature representations.

\subsection{Ablation Study and Analysis}
\paragraph{Does MCL make networks more similar?} With the mutual mimicry by MCL, one may ask if output embeddings of different networks in the cohort would get more similar. To answer this question, we visualize the learned embedding spaces of two ResNet-32 with independent training and MCL, as shown in Fig.~\ref{tsne}. We observe that two networks trained with MCL indeed show more similar feature distributions compared with the baseline. This observation reveals that various networks using MCL can learn more common knowledge from others. Moreover, compared with the independent training, MCL can enable each network to learn a more discriminative embedding space, which benefits the downstream classification performance.

\begin{table}[t]
	\centering
	\resizebox{1.\linewidth}{!}{
		\begin{tabular}{l|c|ccccccccc}  
			\toprule
			Loss&Baseline&\multicolumn{5}{c}{MCL($\times 4$)} \\ 
			\midrule
			$\mathcal{L}^{VCL}$&-&\checkmark&\checkmark&-&-&\checkmark \\
			$\mathcal{L}^{Soft\_VCL}$&-&-&\checkmark&-&-&\checkmark \\
			$\mathcal{L}^{ICL}$&-&-&-&\checkmark &\checkmark&\checkmark \\
			$\mathcal{L}^{Soft\_ICL}$ &-&-&-&-&\checkmark&\checkmark\\
			\midrule
			ResNet-32 &70.91$_{\pm 0.14}$&71.57$_{\pm 0.09}$&73.06$_{\pm 0.15}$&71.92$_{\pm 0.19}$&73.68$_{\pm 0.13}$&\textbf{74.04}$_{\pm 0.07}$\\
			WRN-16-2 &72.55$_{\pm 0.24}$&73.49$_{\pm 0.26}$&74.55$_{\pm 0.11}$&73.89$_{\pm 0.16}$&75.20$_{\pm 0.08}$&\textbf{75.79}$_{\pm 0.07}$ \\
			\bottomrule
	\end{tabular}}
	\caption{Ablation study of loss terms over MCL($\times 4$) on CIFAR-100.}
	\label{ablation}
\end{table}

\begin{table}[t]
	\centering
	\resizebox{1.\linewidth}{!}{
		\begin{tabular}{l|ccccc}  
			\toprule
			$M$&2&3&4&5&6\\ 
			\midrule
			ResNet-32 &72.96$_{\pm 0.28}$&73.74$_{\pm 0.23}$&74.04$_{\pm 0.07}$&\textbf{74.10}$_{\pm 0.10}$&73.98$_{\pm 0.09}$\\
			WRN-16-2&74.56$_{\pm 0.11}$&75.32$_{\pm 0.30}$&75.79$_{\pm 0.07}$&\textbf{75.86}$_{\pm 0.25}$&75.78$_{\pm 0.24}$ \\
			\bottomrule
	\end{tabular}}
	\caption{Ablation study of the number of networks $M$ for MCL on CIFAR-100.}
	\label{num_net}
\end{table}

\paragraph{Ablation study of loss terms in MCL.} As shown in Table~\ref{ablation}, we can observe that each loss term is conducive to the performance gain. Moreover, $\mathcal{L}^{ICL}+\mathcal{L}^{Soft\_ICL}$ outperforms the counterpart of $\mathcal{L}^{VCL}+\mathcal{L}^{Soft\_VCL}$ with an average accuracy gain of 0.64\%. The results verify our claim that ICL and its soft labels are more crucial than the conventional VCL and its soft labels. This is because ICL is more informative than VCL by aggregating cross-network embeddings. Finally, summarizing VCL and ICL into a unified MCL framework can maximize the performance gain for collaborative representation learning. 
\paragraph{Impact of the number of networks $M$.} It is interesting to examine performance gains as the number of networks for MCL increases. As shown in Table~\ref{num_net}, We start from $M=2$ to $M=6$ and find accuracy steadily increases but saturates at $M=5$ on both ResNet-32 and WRN-16-2.

\section{Conclusion}
We propose a simple yet effective Mutual Contrastive Learning method for collaboratively training a cohort of models from the perspective of contrastive representation learning. 
Experimental results show that it can enjoy broad usage for both supervised and self-supervised learning. We hope our work can foster future research to take advantage of collaborative training from multiple networks to enhance supervised or self-supervised representation learning.

{
	\bibliographystyle{aaai22}
	\bibliography{aaai22}
}

\end{document}